%% file: main.tex
\documentclass{article}

\usepackage{arxiv}

\usepackage[utf8]{inputenc}
\usepackage[T1]{fontenc}
\usepackage{upquote}
\usepackage{xurl}
\usepackage{hyperref}
\usepackage{booktabs}
\usepackage{amsfonts}
\usepackage{nicefrac}
\usepackage{microtype}
\usepackage{cleveref}
\usepackage{graphicx}
\usepackage[numbers,sort&compress]{natbib}
\usepackage{doi}

\title{Shorter Is Not Cheaper: A Controlled Comparison of Declarative Animation Specifications and Free-Form Code for LLM-Generated Explanations}

\author{
	Jiantang Huang \\
	\texttt{huang.jiant@northeastern.edu} \\
}

\renewcommand{\shorttitle}{Shorter Is Not Cheaper}

\hypersetup{
pdftitle={Shorter Is Not Cheaper: A Controlled Comparison of Declarative Animation Specifications and Free-Form Code for LLM-Generated Explanations},
pdfsubject={cs.CL, cs.HC},
pdfauthor={Jiantang Huang},
pdfkeywords={generative UI, declarative animation, LLM output representation, latency, reasoning tokens},
}

\begin{document}
\maketitle

\begin{abstract}
Chat assistants increasingly answer conceptual questions with interactive, animated explanations,
generated either as free-form HTML/JavaScript or as declarative specifications that a fixed runtime
plays. The declarative route is widely expected to be cheaper, faster and more reliable, because its
outputs are shorter and constrained. We test this expectation in controlled comparisons whose
hypotheses and analysis plans were frozen before the runs. We designed \emph{ExplainSpec}, a
time-indexed declarative animation format, and an observation contract that makes every generated
explanation programmatically inspectable; validated, pacing-agnostic checkers score 20 explanation
requests from five domains. Three Claude models (360 jobs) and, in a
follow-up, the open-weight Qwen3-14B with thinking enabled and disabled (240 jobs) answered every
request in both formats. For every model the specifications were 25--48\% shorter in answer tokens
(output excluding reasoning), yet with reasoning enabled total output fell significantly only for the
largest Claude model and no model generated significantly faster. The reasons differed. The Claude models reasoned far more
before writing specifications (exploratory; for the two larger models, from 15--20\% to 85--88\% of
generations). Qwen3-14B reasoned about as much in both formats (a pre-registered
test of this shift was not supported), and its shorter answers saved little against a reasoning trace
more than twice as long. Nor were specifications more reliable: free-form pages were valid after
one repair round in 95--100\% of jobs for every model, whereas Qwen3-14B produced valid specifications
in only 33\% of jobs and passed all semantic tests in 17\% (HTML/JavaScript: 62\%). For reasoning
models, a shorter format alone therefore does not make generation cheaper, faster or more reliable,
and a format learned in context can cost reliability.
\end{abstract}


\section{Introduction}
\label{sec:intro}

Interactive, visual answers have become a standard capability of chat assistants: deployed systems
explain mathematics and science with parameterised visuals and simulations inside the
conversation~\cite{openai2026mathscience,anthropic2026visuals,google2026interactive,openai2026intelligentui}.
Because such explanations are generated on demand, while the user waits, their generation cost and
latency matter as much as their quality. Two ways of producing them coexist. The model can write a
complete, self-contained HTML/JavaScript page, which is maximally expressive but long: complete
generated pages can take a minute or more~\cite{leviathan2026generative}. Alternatively, the model can
emit a compact declarative description that a trusted, prebuilt runtime renders, the approach taken
by generative-UI catalogs such as A2UI, json-render and OpenUI~\cite{google2025a2ui,vercel2026jsonrender,thesys2026openui}
and, according to its announcement, by ChatGPT's Intelligent UI~\cite{openai2026intelligentui}.

The case for the declarative route rests on two intuitions. First, because decoding output tokens
dominates generation latency~\cite{openai2026latency}, a shorter artifact should arrive sooner and cost
less. Second, a constrained format with a validating runtime should fail less often than free-form
code. Neither intuition has been tested under controlled conditions for animated explanations, and the
indirect evidence points in different directions: a structured trace pipeline for algorithm videos
improved correctness but was reported to consume more tokens than direct code~\cite{liao2026algogen};
a harness that generates component-constrained UI code reported fewer output tokens than raw HTML
without controlling for quality~\cite{mindlab2026macaron}; format restrictions can interact with
reasoning~\cite{tam2024speak,kurt2024say,fan2026capacity}; and frontier models already produce running
interactive pages most of the time, although often with wrong content~\cite{chen2026interactscience}.

Testing these intuitions fairly is harder than it appears. The two formats must convey the same
content, so that differences can be attributed to the representation rather than to the task;
correctness must be checked for animated, parameterised output regardless of how the model paces its
animation; and the generating model must be isolated from the experimenter's tools and instructions.
We meet these requirements as follows. Both formats implement one programmatic observation contract,
so that the same checkers apply to both. The checkers compare state traces with ground truth
independently of pacing, and they were validated against independent reference implementations and by
mutation testing before any model output existed. Every generation runs in an isolated process, and
the hypotheses, measures and analysis plan were written down and frozen before the main run. Within
this design, three Claude models of different sizes answered 20 requests in physics, algorithms,
network protocols, machine learning and probability, three times in each format, with one repair round
for invalid artifacts. In a follow-up with its own frozen plan, we repeated the study with an
open-weight reasoning model from another family, Qwen3-14B, run locally with thinking enabled and, as an
ablation, disabled.

The results do not support either intuition for the models we tested (\cref{fig:ratios},
\cref{sec:qwen}). With reasoning enabled, the specifications are 25--48\% shorter in answer tokens for
all four models, yet generation is not significantly faster for any of them, and total output falls
significantly only for the largest Claude model. Why the saving vanishes differs between the families:
the Claude models reason far more before writing a specification, whereas Qwen3-14B, for which a
pre-registered test of this shift was not supported, reasons at length in both formats, so that its much
shorter answer changes the total little. Nor is the constrained format more reliable. Free-form pages are
valid after at most one repair round in 95--100\% of jobs for every model, while specifications are less
often valid; for Qwen3-14B they pass all semantic tests in only 17\% of jobs, against 62\% for
HTML/JavaScript. The central lesson is that output
compression and generation savings are different quantities: what a reasoning model writes is only
part of what it generates, and a format learned in context can make the explanation itself less
reliable.

\paragraph{Contributions.}
(1) Evidence from controlled comparisons with four models from two families that a declarative
animation format shortens the answer but, with reasoning enabled, neither speeds up generation nor
improves reliability, and that the reasons differ between the model families
(\cref{sec:results,sec:qwen}).
(2) ExplainSpec, a time-indexed declarative animation format with runtime-provided playback and
parameter controls, together with an observation contract that makes free-form and declarative
explanations equally inspectable (\cref{sec:method}).
(3) A benchmark of 20 explanation tasks with pacing-agnostic checkers, validated by 40 independent
reference implementations and mutation testing, and a taxonomy of the errors models make in an
in-context-learned declarative language (\cref{sec:setup,sec:results}).

\input{related_work}

\section{ExplainSpec and the observation contract}
\label{sec:method}

\paragraph{Format.}
An ExplainSpec document is a JSON object whose main parts are (\cref{fig:spec}) \texttt{params}
(user-adjustable parameters with ranges), \texttt{constants}, \texttt{state} (initial values),
\texttt{steps} (the timeline), \texttt{derived} (values recomputed every frame), \texttt{expose} (the
variables reported to an inspector) and \texttt{objects} (what is drawn), plus an optional title and
canvas size. The timeline is a sequence of step instances. Entering an instance applies literal
assignments (\texttt{set}) and then expressions (\texttt{compute}); the new values hold for
\texttt{dur} seconds. Steps can be repeated (\texttt{repeat}), looped while a condition holds
(\texttt{while}) or grouped, so that the runtime can compute a whole algorithm trace from the
parameters. The timeline variables \texttt{time}, \texttt{step} (counted from 0) and the in-step
progress \texttt{p} allow continuous motion, for example binding a physical time to \texttt{p} and
positions to closed-form expressions. Seven object types are available: \texttt{rect},
\texttt{circle}, \texttt{line}, \texttt{text}, \texttt{path} (a curve sampled from expressions),
\texttt{array} (a row of cells with highlighting) and \texttt{repeat} (instantiating child objects for
an index range). Any object property can be a literal or an expression introduced by \texttt{=}.

\paragraph{Expression language.}
Expressions look like JavaScript but are parsed and interpreted by the runtime instead of being
executed as code. They support numbers, strings, arrays, object literals, indexing, member access,
arithmetic, comparison, logical and conditional operators, and a whitelist of pure functions
(\texttt{sin}, \texttt{floor}, \texttt{clamp}, \texttt{lerp}, \texttt{len}, \texttt{push},
\texttt{swap}, \texttt{setAt}, \texttt{slice}, \texttt{contains}, \dots). There are no statements,
loops inside expressions, user-defined functions or global objects; iteration is expressed with
\texttt{repeat} and \texttt{while} steps. The validator checks every expression for syntax errors and
unknown variables before playback, and the timeline is capped at 2{,}000 step instances.

\paragraph{Runtime.}
The runtime (about 600 lines of JavaScript) validates the document, compiles the timeline for the
current parameter values, renders the scene as SVG, and provides play/pause, a scrubber, step buttons
and one slider per parameter. Changing a parameter recompiles the timeline.

\paragraph{Observation contract.}
To evaluate free-form and declarative artifacts with the same checkers, every artifact must expose
\texttt{window.\_\_explain} with a \texttt{duration}, the \texttt{params} with their current values, a
deterministic \texttt{seek(t)}, a \texttt{getState()} returning the task's state variables at the
current time, and \texttt{setParam(name, value)}. The ExplainSpec runtime implements the contract
automatically from \texttt{expose}, deriving the exposed state and the drawing from the same
variables; a free-form page has to implement the contract itself. Each distinct discrete state must
remain visible for at least 0.3\,s, so that dense sampling cannot miss it.

\begin{figure}[t]
\centering
\begin{minipage}{0.96\linewidth}
\footnotesize
\begin{verbatim}
{ "params": { "a0": {"min": 1, "max": 200, "default": 84, "step": 1},
              "b0": {"min": 1, "max": 200, "default": 36, "step": 1} },
  "state":  { "done": false },
  "steps": [
    { "dur": 1.5, "compute": {"a": "a0", "b": "b0", "done": "b0 == 0"} },
    { "while": "b != 0", "dur": 1.5,
      "compute": {"r": "a % b", "a": "b", "b": "r", "done": "r == 0"} },
    { "dur": 2, "label": "=\"b = 0, so gcd = \" + a" } ],
  "expose": ["a", "b", "done"],
  "objects": [
    { "type": "rect", "x": 100, "y": 150, "w": "=min(600, a * 3)", "h": 40,
      "text": "='a = ' + a" },
    { "type": "rect", "x": 100, "y": 220, "w": "=min(600, b * 3)", "h": 40,
      "text": "='b = ' + b", "visible": "=b > 0" } ] }
\end{verbatim}
\end{minipage}
\caption{An abridged ExplainSpec document (one of the two worked examples given to the models):
Euclid's algorithm, computed by the runtime from the parameters with a \texttt{while} step.}
\label{fig:spec}
\end{figure}

\section{Experimental design}
\label{sec:setup}

The design implements the three requirements stated in the introduction: the tasks and prompts give
both formats the same content, the evaluation checks correctness independently of pacing and was
validated before use, and every generation is isolated from the experimenter's environment.

\paragraph{Tasks.}
We wrote 20 explanation requests, four per domain (\cref{tab:tasks}). Each request states the
question, the parameters the user can change (range, default, step), the state variables the
explanation must expose (names, units, meanings) and precise rules wherever the subject admits
several conventions, for example the Lomuto partition scheme, the tie-breaking rule of $k$-means or
the congestion-window update of TCP. Every task is tested at 2--3 parameter settings chosen so that an
explanation that ignores the parameter fails. Content that changes over time is the substance of
every task, the condition under which animation is most plausibly helpful, although the evidence that
animation aids learning more than static graphics is not encouraging~\cite{tversky2002animation}.

\begin{table}[t]
\centering
\caption{The 20 tasks. Each task is tested at its default and 1--2 further parameter settings
(``Settings'').}
\label{tab:tasks}
\scriptsize
\input{tables/tasks}
\end{table}

\paragraph{Conditions and prompts.}
In the \emph{HTML/JS} condition the model writes one self-contained page (no network access) that
animates the explanation, offers controls and implements the observation contract. In the
\emph{Spec} condition the model writes an ExplainSpec document. Both prompts contain the identical task
brief, the same rules (names and units, 0.3\,s per discrete state, consistency of drawing and state)
and the same two worked examples, for two development tasks that are not part of the evaluation
(uniform circular motion and Euclid's algorithm), each written in the condition's own format. The
HTML/JS examples therefore contain code for the contract and the controls that the model can adapt,
whereas the Spec examples rely on the runtime for both. The HTML/JS prompt additionally contains the
contract and the Spec prompt the format documentation, so that the prompts have similar lengths (for
the TCP handshake task, 12.3k characters for HTML/JS and 11.2k for Spec). Prompts were never revised
in response to outputs of the models under study. While the reference implementations of the
evaluation tasks were being written (see \emph{Validation of the checkers} below), however, the
expression language was extended (object literals and member access, without which the TCP handshake
task could not be expressed) and the format documentation was clarified; the documentation was then
checked on two smoke-test tasks that are not evaluated either. If an artifact is invalid, the model receives one repair prompt with the errors a
developer would see (load errors, uncaught exceptions, console errors, runtime or validator errors,
contract problems and blocked network requests), never the semantic test results; this feedback has
the same format in both conditions.

\paragraph{Models and generation.}
We used three models of different sizes through the Claude Code command-line interface (version
2.1.293): claude-opus-5-5, claude-sonnet-5-5 and claude-haiku-5-5, all with effort \texttt{low}. Every
generation ran in a separate headless process with a neutral system prompt (plus a short fixed
preamble that the interface inserts, identical in every call), no tools, no Model Context Protocol
(MCP) servers, no user settings or plugins beyond the interface's three built-in ones, no persistent
session and an empty working directory. We verified the isolation without relying on the model's
self-report: the interface's initialisation record lists no tools and no servers, and in a positive
control with the user-level instruction file enabled, the same probe call is 989 tokens longer and the
model quotes that file. Each of the 20 tasks $\times$ 2 conditions $\times$ 3 models was generated three
times; we call one such generation, together with its possible repair, a \emph{job} (360 jobs, that is,
180 pairs). Jobs ran in an order shuffled within each repetition, so that the conditions were
interleaved in time, with four jobs running concurrently.

\paragraph{Evaluation.}
A Playwright harness loads every artifact in headless Chrome (fixed viewport, network blocked) and
records the first contentful paint and the time at which the contract becomes callable. The harness
reads the state only through the contract; it does not operate the visible controls. An artifact is
\emph{valid} if it loads without an uncaught exception and the contract is complete. For every tested
parameter setting the harness calls \texttt{setParam}, then samples \texttt{getState()} every 0.05\,s
over the whole timeline (at most 6{,}000 samples), and a task-specific checker compares the trace with a
ground-truth model. Checkers are pacing-agnostic: for discrete processes they compare the sequence of
distinct states (and required intermediate configurations, such as the queue after each
breadth-first-search expansion) with the ground truth; for continuous processes they check the
physical relationships at every sample (e.g.\ $y = v_0 \sin\theta\, t - g t^2/2$ within 1\% of the
maximum height plus 1\,cm) and the coverage of the required time span. A job \emph{passes} if the
final artifact is valid and all tested settings pass. Two descriptive checks complement the tests: a
determinism probe (the same time gives the same state regardless of the order of \texttt{seek} calls)
and a scene-change check (the rendered pixels differ between 20\% and 80\% of the timeline, with the
form controls hidden). Generation timing is taken from the interface's timestamped event stream,
measured from its session-initialisation event: time to first token (first content delta), time to
first answer token (first text delta), and generation latency (until the final result). Time to first
render adds the first contentful paint, and time to first working interaction adds the time until the
contract is callable.

\paragraph{Validation of the checkers.}
Before any model output existed, five independently instructed AI coding agents (Claude-based
sub-agents of the coding assistant used to build the harness) that saw only the task briefs, the
contract and the format documentation, but not the checker code, wrote one reference implementation
per task in each format. All 40 pass. We then corrupted every reference trace systematically
(parameter ignored, a variable perturbed, a discrete value dropped, two values swapped, truncation, a
variable removed, a frozen animation). Of 1{,}430 non-trivial mutants, 1{,}408 (98.5\%) are rejected;
all 22 survivors were inspected and are equivalent with respect to the task rules (for example,
dropping an intermediate state that the rules do not require).

\paragraph{Pre-registration and analysis.}
Hypotheses, measures, exclusion rules and the analysis plan were written down, and all files defining
the experiment (prompts, examples, documentation, tasks, checkers, runtime, harness, generator and
orchestrator) were frozen with SHA-256 hashes before the first main-run generation; the hashes were
unchanged after the run. The pre-registration was internal rather than lodged with an external
registry, and the analysis script was not among the hashed files. \textbf{H1}: the Spec condition needs
fewer output tokens and less generation time (one-sided). \textbf{H2}: the Spec condition is valid at
least as often, with a larger advantage for the smaller model (one-sided). \textbf{H3}: the semantic
pass rates differ (two-sided). \textbf{H4} (descriptive): which tasks the Spec condition can express.
The unit of analysis is the task: for each model we average (rates) or take the median (tokens,
latency) over repetitions and analyse the 20 paired task-level differences with the Wilcoxon
signed-rank test~\cite{wilcoxon1945individual}, using its exact permutation distribution, bootstrap
95\% confidence intervals (CIs) over tasks (10{,}000 resamples)~\cite{efron1979bootstrap}, and Holm's
correction across the three models within each measure~\cite{holm1979simple}. A smoke test on two
non-evaluated tasks, run before the freeze, showed that the models reason even at effort \texttt{low},
and more for specifications; we therefore added answer tokens (output minus reasoning tokens) as a
secondary measure before the freeze. All other analyses of reasoning are exploratory. Generation cost
US\$14.00 for the main run and US\$14.77 including the smoke test and the post-hoc repairs of
\cref{sec:res-robust}, as estimated by the interface at list prices.

\section{Results}
\label{sec:results}

This section reports the main study with the three Claude models; \cref{sec:qwen} reports the follow-up.
All 360 jobs completed without infrastructure failure, and no sampling timed out. \Cref{tab:descriptive}
summarises the outcomes and \cref{tab:tests} the pre-registered tests. We first ask whether the shorter
format makes generation cheaper and faster (\cref{sec:res-eff}) and, because it does not, where the
saved tokens go (\cref{sec:res-think}). We then test the second intuition, that a constrained format is
more reliable (\cref{sec:res-valid}), check that these conclusions survive a correction and relaxed
scoring (\cref{sec:res-robust}), and finally examine which errors the declarative language itself
induces (\cref{sec:res-dsl}).

\begin{table}[t]
\centering
\caption{Outcomes by model and format (60 jobs each). Token and latency columns are medians over first
attempts (latency is generation time); answer tokens exclude reasoning tokens, which the interface
reports as ``thinking tokens''; reasoning jobs: first attempts with any reasoning (of 60). Cost is the
total for the 60 jobs, for first attempts only and including repair rounds.}
\label{tab:descriptive}
\scriptsize
\input{tables/descriptive}
\end{table}

\begin{table}[t]
\centering
\caption{Pre-registered tests on 20 paired tasks. Brackets: 95\% bootstrap CIs. H1 and the secondary
measure use first attempts, H2 and H3 the results after repair. Exact Wilcoxon signed-rank tests,
one-sided in the hypothesised direction for H1, H2 and the secondary measure and two-sided for H3, with
Holm adjustment across the three models; ``--'': all differences are zero, so no test is possible;
pp: percentage points.}
\label{tab:tests}
\scriptsize
\input{tables/tests}
\end{table}

\subsection{The artifact shrinks, but generation does not get cheaper}
\label{sec:res-eff}

\begin{figure}[t]
\centering
\includegraphics[width=\linewidth]{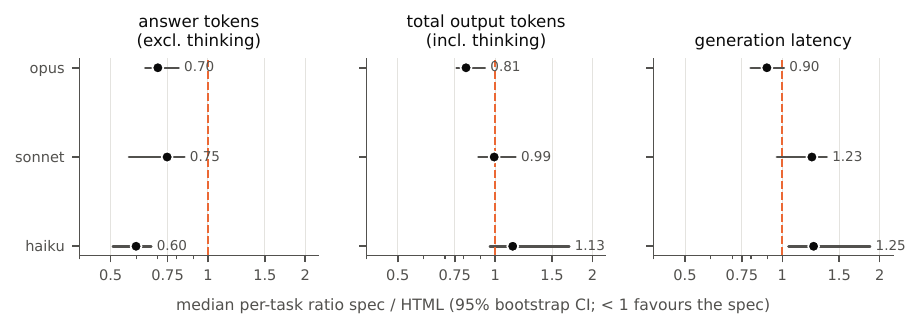}
\caption{Median per-task ratio Spec\,/\,HTML/JS with 95\% bootstrap CIs (log scale; dashed line: no
difference). Specifications are consistently shorter in answer tokens; total output, which includes
reasoning, falls only for Opus, and generation latency is not significantly reduced for any model.}
\label{fig:ratios}
\end{figure}

Specifications are much shorter than HTML pages: the median artifact has 3{,}581 characters against
5{,}868, and answer tokens fall to 0.60--0.75 of the HTML/JS value for every model (\cref{fig:ratios},
left; all Holm-adjusted $p<0.001$). Total output tokens, however, fall only for Opus (0.81, Holm
$p=0.040$ within the measure and $0.080$ when all six H1 tests are adjusted together); for Sonnet they
are unchanged (0.99), and for Haiku the point estimate is 1.13 with a CI that includes 1, with a heavy
tail of long specification generations (90th percentile 14{,}557 vs 8{,}702). Generation latency follows total output (\cref{fig:latency}): it is not significantly reduced
for any model (Opus 0.90, CI [0.80, 1.01]); for Sonnet the point estimate is 1.23 with a CI that
includes 1, and for Haiku it is 1.25 with a CI that excludes 1 in the direction opposite to the
hypothesis (a reverse test was not pre-registered). Rendering adds little (median first contentful
paint 32\,ms vs 40--44\,ms, the latter including the inlined runtime), so the median time to first render
of first-attempt-valid artifacts follows generation latency (HTML/JS vs Spec: 21.1 vs 19.8\,s for Opus,
13.7 vs 18.5\,s for Sonnet, 21.6 vs 28.2\,s for Haiku); including the repair rounds changes these medians by
at most 1\,s, and the contract becomes callable within 0.1\,s of first render. On first attempts the Spec condition is cheaper for Opus (US\$3.89 vs 4.43 for 60 jobs) but
not for Sonnet (2.16 vs 2.13) or Haiku (0.26 vs 0.20); including the repair rounds it is more expensive
for all three models (\cref{tab:descriptive}). H1 is supported only for the total tokens of the largest
model.

\subsection{Where the savings go: reasoning before writing}
\label{sec:res-think}

The difference between answer and total tokens is reasoning (\cref{tab:descriptive,fig:tokens}); the
analyses in this subsection are exploratory. With effort \texttt{low}, Opus and Sonnet reasoned in only 9 and 12 of
60 HTML/JS generations but in 51 and 53 of 60 Spec generations; Haiku reasoned in almost every
generation and roughly doubled its median reasoning in the Spec condition (2{,}454 to 4{,}710 tokens).
The shift is consistent across tasks: specifications drew more reasoning in 17 of 20 tasks for Opus and
15 of 20 for Sonnet, and less in only one task (exact signed-rank test, $p<0.001$ for both). It is strongest
for algorithm tasks, where the share of generations with reasoning rises from 0\% to 100\% for both
models, and weakest for physics tasks (from 25--42\% to 67\%). The reasoning is also visible in the
timing: the median time to the first \emph{answer} token is 0.9 vs 5.3\,s for Opus, 0.9 vs 9.3\,s for
Sonnet and 12.4 vs 22.9\,s for Haiku, whereas the median time to the first token of any kind is
0.7--1.6\,s in both conditions. For Opus, the added reasoning offsets part of the saving (a median of
about 350 additional reasoning tokens against 875 fewer answer tokens per task); for Sonnet and Haiku
it exceeds the saving (870 vs 680 and 2{,}350 vs 1{,}180 tokens). \Cref{sec:discussion} examines two
mechanisms that can produce this shift.

\subsection{Free-form pages were always valid; specifications were not more correct}
\label{sec:res-valid}

\begin{figure}[t]
\centering
\includegraphics[width=\linewidth]{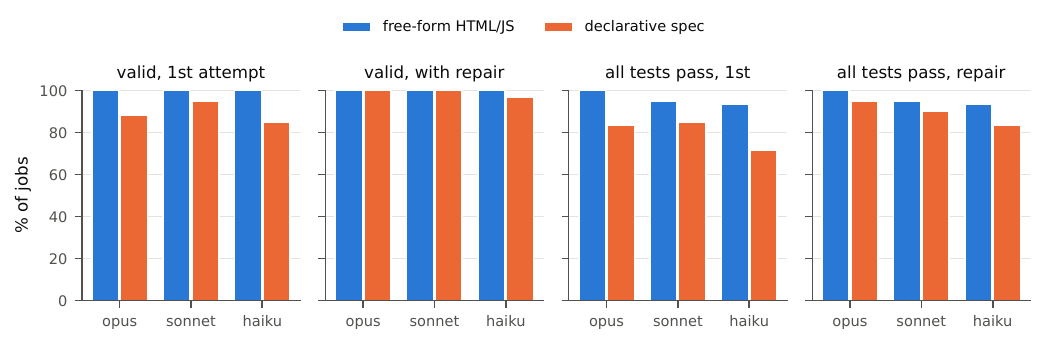}
\caption{Validity and semantic pass rates by model and format, on the first attempt and after the
single repair round (frozen protocol).}
\label{fig:rates}
\end{figure}

The second intuition holds that a constrained format fails less often than free-form code. Free-form
code, however, left little room for improvement: every one of the 180 HTML/JS artifacts was valid on
the first attempt, without a single uncaught exception. Specifications were valid on the first attempt
in 85--95\% of cases; 19 Spec jobs (7 Opus, 3 Sonnet, 9 Haiku) and no HTML/JS job needed the repair
round, after which specifications were valid in 100\% (Opus, Sonnet) and 97\% (Haiku) of cases
(\cref{fig:rates}). The Spec condition was thus equally often valid after repair for the two larger
models and less often for Haiku, and less often valid on the first attempt for all three; H2,
including its prediction of a larger advantage for the smallest model, is not supported. After repair,
all semantic tests were passed by 93--100\% of HTML/JS jobs and 83--95\% of Spec jobs. The differences
($-5$, $-5$ and $-10$ points for Opus, Sonnet and Haiku) have the same direction for all models but
are not significant (exact Holm-adjusted $p=0.75$ for each). Because 14--17 of the 20 tasks are at
ceiling in both conditions, few tasks differ: for Opus and Haiku only three, so that the smallest
attainable two-sided $p$ is 0.25 (0.75 after Holm adjustment); for Sonnet six tasks differ, but in both
directions ($p=0.625$). H3 is therefore inconclusive. On the first attempt the difference is larger: in an exploratory
count, 173 of 180 HTML/JS jobs and 144 of 180 Spec jobs pass, and all 12 tasks whose pass rate differs
when pooled over models favour HTML/JS. Every task was solved by at least one specification (H4),
although the language had been extended before the freeze so that the TCP handshake task could be
expressed (\cref{sec:setup}).
Descriptively, the rendered scene of first-attempt-valid artifacts differed between the 20\% and 80\%
points of the timeline in 97\% (HTML/JS) and 98\% (Spec) of cases, and every one of them passed the
determinism probe.

To see what lies behind these differences, \cref{tab:taxonomy} classifies the 26 jobs that did not
pass, each under its first matching category. Sixteen are classified as strict readings of the briefs:
eight Spec jobs add a leading state with index 0 where the brief starts at 1, which the timeline
variable \texttt{step}, counted from 0, may induce; three HTML/JS jobs reset \texttt{mid} to null
between binary-search probes; and five fall under an ambiguity in the Go-Back-N brief that a
reference-implementation agent had reported before the run (we kept the frozen interpretation). Three
are affected by the harness asymmetry discussed next. Counting only errors beyond these readings and the
asymmetry, one HTML/JS job and ten Spec jobs fail, for example through an incorrectly implemented
partition step or a breadth-first search that stops after two nodes. Per-task results are shown in
\cref{fig:heatmap} and per-task tokens in \cref{fig:tokens} (\cref{app:figures}).

\begin{table}[t]
\centering
\caption{Classification of the 26 jobs that do not pass all semantic tests after repair (frozen
protocol). Each job is counted in the first matching category. In brackets: jobs that still fail after
the correction and with relaxed checkers (\cref{sec:res-robust}), that is, jobs with errors beyond the
strict readings and the harness asymmetry.}
\label{tab:taxonomy}
\small
\input{tables/taxonomy}
\end{table}

\subsection{The conclusions survive a post-hoc correction and relaxed scoring}
\label{sec:res-robust}

Because several of these failures depend on how strictly the briefs are read, and some on how the
harness treats runtime errors, we checked whether either affects the conclusions. A post-run audit
found that the ExplainSpec runtime catches errors raised while a frame is computed or drawn, whereas an
HTML/JS page throws the equivalent error out of \texttt{seek()} and is therefore invalid and repaired.
Five Spec artifacts for the DNS task, which indexed an object with a dynamic key, were thus scored valid
without repair (three with null state, two with undrawn message arrows). We keep the frozen results as
primary and, in a documented deviation, re-evaluated all 379 artifacts with a corrected copy of the
harness that treats such errors like exceptions: every other artifact reproduced its frozen outcome
(validity, passed tests and, for first attempts, failure messages), confirming that the harness is
deterministic, and the five passed after repair. An exploratory analysis additionally relaxed the
three strict readings for both formats. Neither the correction nor the relaxation changes the direction
of any result (\cref{tab:sensitivity}); under relaxed scoring the Spec condition is within one or two jobs of HTML/JS
for Opus and Sonnet but remains 12 points less often correct for Haiku.

\begin{table}[t]
\centering
\caption{Jobs passing all semantic tests after repair (\%, HTML/JS vs Spec) under the frozen protocol,
after treating runtime errors like exceptions (corrected), and additionally with relaxed checkers
(exploratory).}
\label{tab:sensitivity}
\small
\input{tables/sensitivity}
\end{table}

\subsection{Errors in an in-context-learned declarative language}
\label{sec:res-dsl}

The lower first-attempt validity of specifications has a visible source in the errors that the
validator and the runtime report: of all Spec artifacts, 26 triggered such an error
(\cref{tab:usability}). The most frequent errors carry JavaScript habits into a language that only
looks like JavaScript: indexing an object with a dynamic key (\texttt{px[msg.from]}), calling array
functions on a state that was initialised to null, or placing the object-property prefix \texttt{=}
inside fields that are always expressions. Free-form JavaScript, which the models have seen far more
often, produced no uncaught exception at all.

\begin{table}[t]
\centering
\caption{Validator and runtime errors in Spec artifacts (all attempts, first error per artifact).}
\label{tab:usability}
\small
\input{tables/usability}
\end{table}

\section{Follow-up with an open-weight reasoning model}
\label{sec:qwen}

\paragraph{Design.}
To test whether the results depend on the model family, and to run the condition that the Claude
command-line interface did not allow, reasoning disabled, we repeated the study with
Qwen3-14B~\cite{yang2025qwen3}, an open-weight model whose thinking can be enabled or disabled per
request. The model (14.8B parameters, 4-bit quantisation Q4\_K\_M) was served locally by Ollama 0.34.3
on one consumer GPU with 16\,GB of memory, entirely on the GPU, with a context of 32{,}768 tokens and at
most 24{,}576 generated tokens per call. Tasks, prompts, worked examples, format documentation, contract,
checkers, artifact extraction and repair round were the frozen files of the main study; the generator
sent one request per attempt with the same neutral system prompt, and the corrected harness of
\cref{sec:res-robust} scored the artifacts from the start. Each mode, thinking enabled and disabled,
covered 20 tasks $\times$ 2 formats $\times$ 3 repetitions (240 jobs in total), generated one at a time
with the sampling settings that the model's documentation recommends for each mode and a fixed seed per
attempt. Reasoning tokens were counted from the streamed thinking chunks, whose number differed from the
server's token count by less than 1\%, and latency was measured from the request, with the model already
loaded; it includes prompt processing, which the server partly skipped for repeated prompts (latency
without it gives the same ratio). Because Qwen3 always reasons when thinking is enabled, the main study's measure of how often a
model reasons is uninformative here; the measure is the number of reasoning tokens.

\paragraph{Pre-registration.}
An addendum to the pre-registration, written after a plumbing pilot on the two smoke-test tasks and
frozen with SHA-256 hashes before the first generation on an evaluation task, made the reasoning shift
observed for the Claude models the primary, confirmatory hypothesis (Q1: more reasoning tokens for Spec,
one-sided). It repeated the main study's hypotheses: shorter answers (Q2a), fewer output tokens and
lower latency (Q2b, Q2c; H1), higher validity (Q3; H2) and a difference in correctness (Q4; H3), with the
same task-level exact tests and Holm's correction across the six tests; the runs with thinking disabled
form an exploratory ablation. The pilot had shown more reasoning for Spec on both smoke tasks, which the
addendum discloses; nothing was changed after it. One deviation occurred: a generated page that hung
while loading made the orchestrator raise an exception instead of recording the page as invalid. After a
fix, the job was completed from its saved generation rather than generated anew, so that its actual
outcome (invalid) was kept.

\paragraph{Results with thinking enabled.}
\Cref{tab:qwen,tab:qwentests} summarise the results. The reasoning shift did not replicate: Qwen3-14B
spent a median of 3{,}370 reasoning tokens on HTML/JS and 3{,}686 on specifications (median per-task
ratio 1.07 [0.95, 1.33], Holm $p=0.29$), and specifications drew more reasoning in only 11 of 20 tasks.
In an exploratory breakdown, the only clear shift was for algorithm tasks (median ratio 1.57), the
domain with the strongest shift for the Claude models, whereas protocol tasks drew less reasoning for
specifications (0.76). The answer itself was again much shorter for specifications (ratio 0.52), but it
made up at most a third of the output, so that total output (0.87 [0.76, 1.04]) and latency (0.89 [0.77,
1.11]; medians 104 vs 95\,s) were not significantly reduced. The specification format was, however, far
less reliable than for the Claude models: 13\% of first attempts and 33\% of jobs after repair were
valid, against 85\% and 95\% for HTML/JS, and 17\% against 62\% of jobs passed all semantic tests (Holm
$p=0.004$); under the same corrected harness, the Claude models' specifications were valid in 83--90\% of
first attempts. The failing specifications show error types also seen for the Claude models (\cref{tab:usability}), at
a much higher rate: an \texttt{=} prefix inside fields that are always expressions accounts for 39 of the 93 Spec artifacts
with an error, followed by invalid JSON and constructs from JavaScript or Python that the expression
language lacks, such as \texttt{Math.floor}, lambdas, list comprehensions or \texttt{\^{}} for powers.
Most failing HTML/JS jobs produced pages that loaded but showed wrong content. Because most
specifications were invalid on the first attempt, the token and latency comparisons for this model
contrast mostly valid HTML/JS with mostly invalid specifications.

\paragraph{Ablation with thinking disabled.}
Without thinking, generation was much faster (medians 29\,s for HTML/JS and 18\,s for specifications),
and here the shorter format did pay: answer tokens fell to 0.64 and latency to 0.65 of the HTML/JS value
(both $p<0.001$, exploratory). Specifications, however, were rarely usable: 32\% were valid after repair
(HTML/JS: 97\%) and 3\% passed all semantic tests (HTML/JS: 40\%); one specification reached the token
limit by repeating the same drawing object. Thinking raised the share of jobs passing all tests by 22
percentage points for HTML/JS and by 13 for specifications; the difference between the formats ($-8$
points [$-32$, +17], $p=0.61$) gives no indication that specifications depend more on reasoning than
free-form code.

\begin{table}[t]
\centering
\caption{Qwen3-14B follow-up: outcomes by mode and format (60 jobs each), scored with the corrected
harness. Token and latency columns are medians over first attempts; reasoning tokens are counted from
the streamed thinking; latency is measured on one local GPU.}
\label{tab:qwen}
\scriptsize
\input{tables/qwen_descriptive}
\end{table}

\begin{table}[t]
\centering
\caption{Qwen3-14B follow-up: pre-registered tests with thinking enabled (one Holm family of six tests)
and the exploratory ablation. Effects are median per-task ratios Spec/HTML/JS for tokens and latency and
mean differences in percentage points (pp) for rates, with 95\% bootstrap CIs; exact Wilcoxon
signed-rank tests on 20 paired tasks, one-sided in the stated direction.}
\label{tab:qwentests}
\scriptsize
\input{tables/qwen_tests}
\end{table}

\section{Discussion}
\label{sec:discussion}

\paragraph{Why compression did not pay.}
The efficiency argument for declarative output assumes that the model's work is proportional to the
length of what it writes. For the reasoning models we tested, this assumption did not hold, in two
different ways. The Claude models wrote shorter artifacts but reasoned longer before writing them, and
for two of three models the added reasoning exceeded the saving. Qwen3-14B reasoned about as much for
both formats, but its reasoning was more than twice as long as its answer, so that halving the answer
shortened generation only slightly; only with thinking disabled did the shorter format make generation
significantly faster, and then specifications were rarely correct. What a reasoning model writes is only
part of what it generates, and shortening that part has limited leverage.

Why the Claude models but not Qwen3-14B reasoned more for specifications remains open. It is not simply
that Qwen3-14B always reasons: Haiku also reasoned in almost every generation and still roughly doubled
its reasoning for specifications. Two mechanisms can produce such a shift, and both may contribute.
\emph{Unfamiliarity}: because the format was learned only from the prompt, the model may need to reason
about how to use it. \emph{Encoding and planning}: a specification must fix the sequence of states and
the expressions that compute them before anything is drawn, in a language without loops inside
expressions or user-defined functions, so that iterative computations have to be unrolled or recast as
\texttt{while} steps; a program can compute the same values at run time with familiar constructs. The
domain pattern is consistent with the second account: for the Claude models the shift is largest for
algorithm tasks, whose traces must be encoded step by step, and smallest for physics tasks with
closed-form solutions, and algorithm tasks were also the only domain with a clear shift for Qwen3-14B.
The two accounts also make different predictions: unfamiliarity predicts that the overhead shrinks as
the model becomes familiar with the format, for example with more examples or after fine-tuning,
whereas encoding and planning predict that it persists. The Claude interface reported reasoning token
counts but not reasoning text; Qwen3-14B's reasoning text was saved and permits a direct analysis of
what the model reasons about, which we leave to future work. Disabling reasoning shows what it buys:
for Qwen3-14B it lowered correctness in both formats by similar amounts, so we found no indication that
specifications depend more on reasoning than free-form code. For the Claude models this condition
remains untested, because no switch of the command-line interface that we tested turned reasoning off.

\paragraph{Implications for declarative generative UI.}
In our setting, a shorter format learned in context did not make generation significantly faster for
any model with reasoning enabled and was cheaper only for Opus on first attempts, and the models did not
need it to produce artifacts that run: free-form pages were valid after repair in 95--100\% of jobs for
every model. A format learned only from the prompt can, moreover, cost reliability: the Claude models
produced valid specifications in 83--90\% of first attempts, Qwen3-14B in 13\%, in line with the
observation that small models handle an unseen component catalog poorly~\cite{yang2026frontier}. The case for declarative
output therefore has to rest on other properties, several of which production systems already exploit. Models can be
trained to compose responses from a component library, as reported for Intelligent
UI~\cite{openai2026intelligentui}, or fine-tuned on a catalog, as shown for A2UI
generators~\cite{yang2026frontier}, which removes unfamiliarity; a static format description can be
cached; specifications can be rendered while they stream; and a trusted runtime offers security and
consistency that free-form code does not. The first three are hypotheses that the present design can
test directly.

\paragraph{Design lessons for model-facing domain-specific languages.}
Beyond the comparison itself, the errors in \cref{sec:res-dsl} suggest three inexpensive changes to
such languages: adopt JavaScript semantics wherever a construct looks like JavaScript (dynamic indexing
of objects, null-tolerant array functions); avoid syntactic markers whose meaning depends on the field
(the \texttt{=} prefix, which caused 4 of the 26 Claude errors and was the most frequent error of
Qwen3-14B); and surface runtime errors loudly instead of degrading gracefully, since
graceful degradation hid five failures from both the repair loop and our own first evaluation.

\paragraph{Scope and limitations.}
With two model families, 20 tasks and three repetitions per cell, the study resolves large effects such
as the shifts in tokens and reasoning, but not differences of 5--10 points in correctness, whose
direction is consistent across the Claude models. The open-weight model was a single 4-bit quantised
14B model run locally, so its latencies reflect one consumer GPU, and its token comparisons involve
mostly invalid specifications. Besides the output format, the conditions differ in ways that
any format comparison of this kind entails, and each can contribute to the reasoning shift: the
specification format was learned only from the prompt, the HTML/JS examples contained code for the
contract and the controls that could be adapted, the expression language has no loops inside
expressions, and all Claude generations used one effort level. Each suggests an ablation that our
harness supports without changes to the evaluation: specification prompts with matched demonstrations,
HTML/JS that calls a provided runtime library, an extended expression language, further effort levels
and model families, and reasoning disabled for the Claude models through the API. The checkers verify the exposed state and its consistency with the
rules rather than visual quality or pedagogical value; in the HTML/JS condition the contract is
model-written code that could in principle diverge from the drawing, and the harness does not operate
the visible controls. We saved mid-timeline screenshots of all 358 artifacts that loaded with a working contract,
which would allow a visual rating, although the runtime's recognisable appearance limits blinding.
Because the HTML/JS condition also had to generate its own controls, its output is longer than the
explanation alone requires; the token comparison therefore favours the specification, which makes the
absence of a clear saving more notable. No condition used streaming rendering. The
reference implementations were written by agents of the same model family as the Claude generators, and
several failures are strict readings of explicit rules in briefs we wrote; we report them separately
(\cref{tab:taxonomy,tab:sensitivity}).

\section{Conclusion}
\label{sec:conclusion}

We compared declarative animation specifications with free-form HTML/JavaScript for LLM-generated
interactive explanations in controlled experiments with frozen protocols, validated checkers and four
models from two families. Neither intuition behind declarative output held for reasoning models. A
specification shortened the answer but not generation, because the Claude models reasoned more before
writing it and Qwen3-14B's answer was too small a part of its output to matter; only without reasoning
did the shorter format pay off in speed, and then specifications were rarely correct. A format learned
in context was also less reliable than free-form code, markedly so for the smaller open-weight model.
Output compression is therefore not the same as generation savings. Future work should test formats
that models were trained on, analyse what models reason about when they write a specification, measure
streaming and caching, and rate visual quality.

\section*{Use of AI assistance}
The experimental harness, the reference implementations and parts of this manuscript were developed
with the assistance of an AI coding assistant.

\bibliographystyle{unsrtnat}
\bibliography{references}

\appendix
\section{Additional figures}
\label{app:figures}

\begin{figure}[h]
\centering
\begin{minipage}[b]{0.53\linewidth}
\centering
\includegraphics[width=\linewidth]{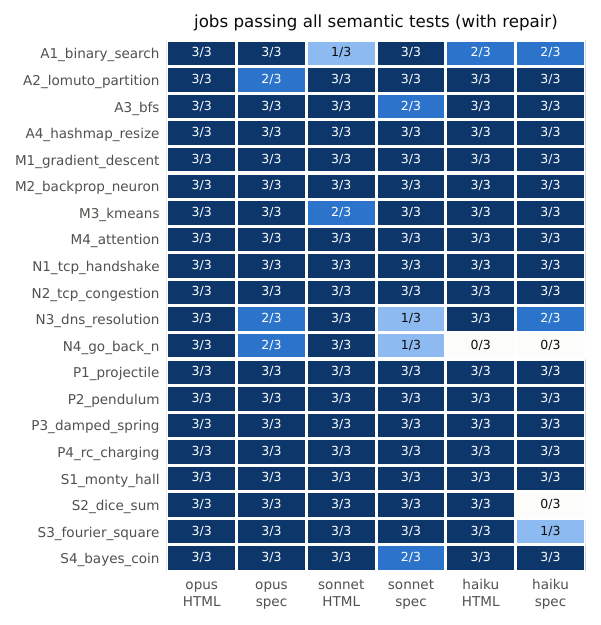}
\caption{Jobs passing all semantic tests after repair, per task, model and format (frozen protocol).
Darker cells: more of the three repetitions pass.}
\label{fig:heatmap}
\end{minipage}\hfill
\begin{minipage}[b]{0.43\linewidth}
\centering
\includegraphics[width=\linewidth]{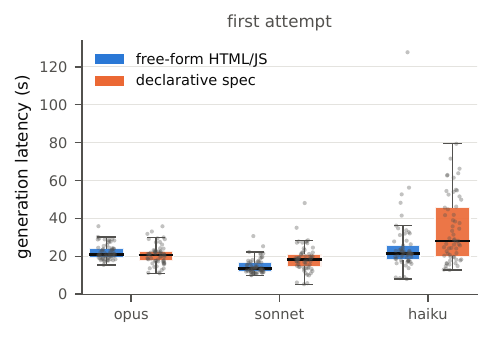}
\caption{Generation latency of first attempts (boxes: interquartile range; dots: individual jobs). The
long upper tail of Haiku's specification generations coincides with its longer reasoning
(\cref{fig:tokens}).}
\label{fig:latency}
\end{minipage}
\end{figure}

\begin{figure}[h]
\centering
\includegraphics[width=\linewidth]{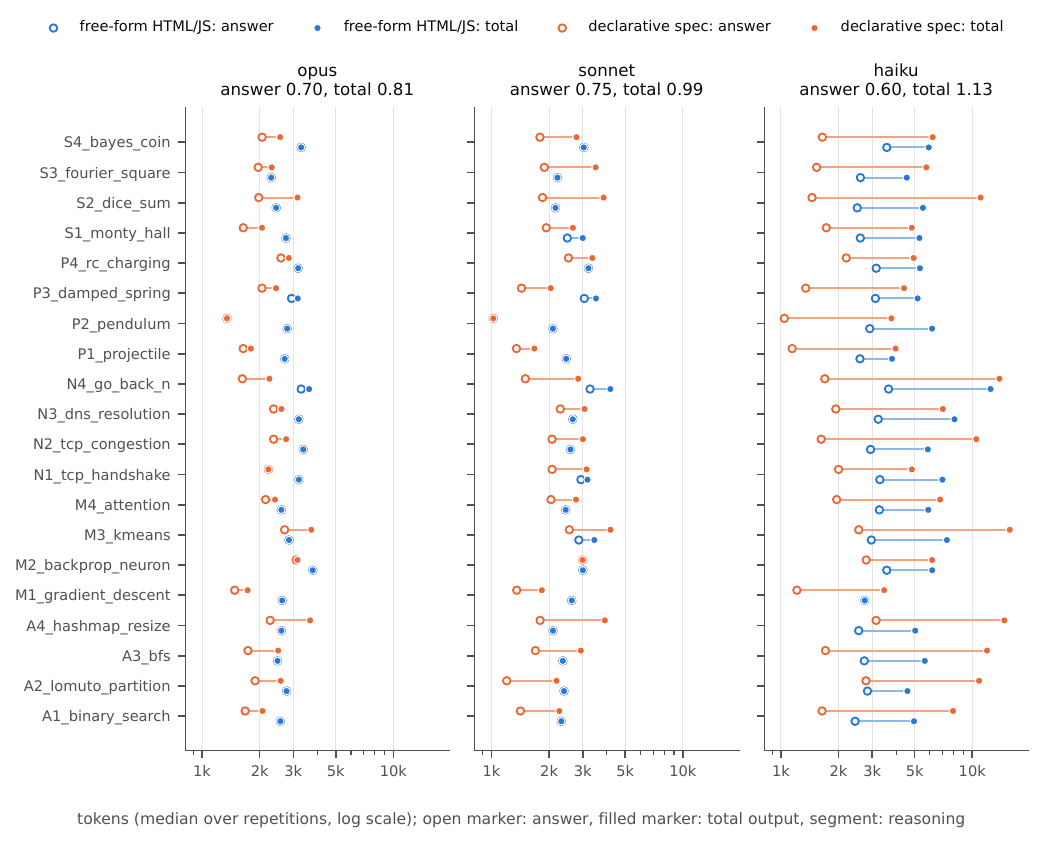}
\caption{Answer and total output tokens per task (medians over repetitions of first attempts). Each
segment runs from the answer tokens (open marker) to the total output (filled marker), so its length is
the reasoning; a lone filled marker means no reasoning. Panel titles give the median per-task ratio
Spec/HTML/JS for answer and total tokens. Specifications are shorter in answer tokens for almost every
task, but with reasoning included they are no longer consistently shorter.}
\label{fig:tokens}
\end{figure}

\end{document}

%% file: related_work.tex
\section{Related Work}
\label{sec:related}

\paragraph{Generative UI: free-form code versus declarative catalogs.}
Interactive answers are now produced in two ways. Systems that let the model write complete
HTML/CSS/JavaScript achieve high quality, but a page can take a minute or more to generate, and the
pipeline relies on post-processors to fix common errors~\cite{leviathan2026generative}; Gemini's
generative interfaces follow this route. Deployed assistants also offer in-chat visuals and
simulations~\cite{anthropic2026visuals,google2026interactive}; according to its documentation, Claude
builds them with HTML~\cite{anthropic2026helpvisuals}, whereas the format of the Gemini app's
simulations is not publicly documented.
The second route constrains the model to a catalog of declarative components that a trusted client
renders, as in A2UI, json-render and OpenUI~\cite{google2025a2ui,vercel2026jsonrender,thesys2026openui},
and, according to its announcement, in ChatGPT's Intelligent UI, whose representation is not
public~\cite{openai2026intelligentui}. Three observations motivate a controlled test. First, the
efficiency evidence for compact UI formats is indirect: a harness that generates component-constrained
UI code reported fewer output tokens than raw HTML without controlling for
quality~\citep[Sec.~3.1.1]{mindlab2026macaron}. Second, small models struggle with a catalog they have not
been trained on: base models with 2B and 4B parameters produce valid A2UI for only 3\% and 62\% of
requests, whereas a fine-tuned 4B model approaches the quality of a frontier model at a fraction of
its cost~\cite{yang2026frontier}. Third, existing catalogs describe interface widgets and playback of
pre-rendered media~\cite{google2026catalog}, and json-render's video renderer turns JSON timelines of
clips into video compositions~\cite{vercel2026jsonrender}; none combines time-indexed animation with
user-adjustable parameters from which the runtime recomputes it, which is what interactive conceptual
explanations require.

\paragraph{LLM-generated explanatory animations.}
Systems that do generate explanatory animation mostly have the model write code: agents write Manim
programs for educational videos and spend 18--26 times the tokens of direct code
generation~\cite{chen2026code2video}, and the InteractScience benchmark finds that self-contained
HTML/JavaScript demonstrations often run while their science is wrong~\cite{chen2026interactscience}.
Several lines come closer to a declarative design. Keyframer lets a model emit CSS keyframes that the
browser plays~\cite{tseng2025keyframer}. DataMagic's multi-agent pipeline writes a declarative
data-video specification that executes more reliably than direct code, although it took about three
times as long~\cite{xie2026datamagic}, and spec-then-compile pipelines with specification repair have
been applied to interactive what-if interfaces, whose direct one-shot generation often fails to
render~\cite{gathani2026bridging}. ALGOGEN is the closest comparison: the model writes a tracker
program whose schema-checked traces drive fixed renderers, which made algorithm videos more correct
but, as reported, consumed more tokens than direct code~\cite{liao2026algogen}. These systems produce
offline video, standalone animations or static interfaces, and none measures generation latency inside
a conversation. The vocabulary itself is not new either: interpreted animation command streams date
back to algorithm animation~\cite{stasko1997using}, and visualisation grammars already compose timed
transitions~\cite{kim2021gemini,zong2023animated}. Our contribution is therefore not the vocabulary but
the controlled comparison of what it costs a model to emit such a specification instead of code.

\paragraph{Output format, reasoning and latency.}
Because decoding dominates latency, shorter outputs should be faster~\cite{openai2026latency}. Whether
a format changes how a model reasons is less clear. Restricting answers to structured formats can
lower reasoning accuracy~\cite{tam2024speak}, but a vendor analysis attributes much of this effect to
prompt and parser artifacts~\cite{kurt2024say}, and it concentrates in models near their
capacity~\cite{fan2026capacity}, which also reports how many tokens a non-reasoning model writes under
different formats; conversely, code is a more effective action format than JSON when steps must be
composed~\cite{wang2024executable}. For reasoning models answering benchmark questions, the number of
thinking tokens tracks question difficulty rather than whether the answer must follow a structured
format~\cite{lee2026format}. To our knowledge, no study has measured how the representation of a
generated artifact changes this hidden reasoning, and with it latency and cost, or compared a
time-indexed declarative animation format with free-form code on identical tasks while measuring
artifact tokens, reasoning tokens, latency and correctness. This is the gap our study addresses.

%% file: tables/tasks.tex
\begin{tabular}{@{}llp{3.0cm}p{4.6cm}c@{}}
\toprule
Domain & Task & Parameters & Required state variables & Settings \\
\midrule
Physics & Projectile motion & \texttt{v0, theta} & \texttt{t, x, y} & 3 \\
Physics & Simple pendulum & \texttt{L, theta0} & \texttt{t, angle} & 3 \\
Physics & Damped spring-mass oscillator & \texttt{k, c} & \texttt{t, x} & 3 \\
Physics & RC circuit charging & \texttt{R, C} & \texttt{t, v\_c, i} & 3 \\
Algorithms & Binary search & \texttt{target} & \texttt{lo, hi, mid, found} & 3 \\
Algorithms & Quicksort partition (Lomuto) & \texttt{variant} & \texttt{arr, j, pivot} & 3 \\
Algorithms & Breadth-first search & \texttt{start} & \texttt{current, queue, discovered} & 3 \\
Algorithms & Hash table resizing & \texttt{variant} & \texttt{capacity, size, buckets} & 3 \\
Protocols & TCP three-way handshake & \texttt{client\_isn, server\_isn} & \texttt{client\_state, server\_state, last\_segment} & 3 \\
Protocols & TCP slow start and congestion avoidance & \texttt{ssthresh0, loss\_round} & \texttt{round, cwnd, ssthresh} & 3 \\
Protocols & DNS iterative resolution & \texttt{tld\_cached} & \texttt{msg, answered} & 2 \\
Protocols & Go-Back-N sliding window & \texttt{window, lost} & \texttt{round, base, sent, delivered} & 3 \\
ML & Gradient descent & \texttt{lr, w0} & \texttt{iter, w, loss} & 3 \\
ML & Backpropagation through one neuron & \texttt{x, w, b, target} & \texttt{phase, z, y, L, dL\_dy, dL\_dz, dL\_dw, dL\_db} & 3 \\
ML & k-means clustering & \texttt{init} & \texttt{iter, centroids, assign} & 3 \\
ML & Scaled dot-product attention & \texttt{q1} & \texttt{stage, scores, scaled, weights, output} & 3 \\
Prob./math & Monty Hall problem & \texttt{pick} & \texttt{car, opened, stay\_wins, switch\_wins} & 3 \\
Prob./math & Distribution of the sum of dice & \texttt{n} & \texttt{k, dist} & 3 \\
Prob./math & Fourier series of a square wave & \texttt{terms} & \texttt{k, value\_at\_pi\_2, amplitude} & 3 \\
Prob./math & Bayesian updating of a coin & \texttt{a0, b0} & \texttt{n, alpha, beta, posterior\_mean} & 3 \\
\bottomrule
\end{tabular}

%% file: tables/descriptive.tex
\begin{tabular}{@{}llrrrrrrrrrrr@{}}
\toprule
 & & \multicolumn{2}{c}{Valid (\%)} & \multicolumn{2}{c}{All tests pass (\%)} & \multicolumn{3}{c}{Output tokens (median)} & Reasoning & Latency & \multicolumn{2}{c}{Cost (\${,} 60 jobs)} \\
\cmidrule(lr){3-4}\cmidrule(lr){5-6}\cmidrule(lr){7-9}\cmidrule(lr){12-13}
Model & Format & 1st & repair & 1st & repair & total & answer & reasoning & jobs & (s) & 1st & repair \\
\midrule
Opus & HTML/JS & 100 & 100 & 100 & 100 & 2{,}806 & 2{,}795 & 0 & 9 & 21.1 & 4.43 & 4.43 \\
 & Spec & 88 & 100 & 83 & 95 & 2{,}472 & 2{,}043 & 402 & 51 & 20.7 & 3.89 & 4.63 \\
\addlinespace
Sonnet & HTML/JS & 100 & 100 & 95 & 95 & 2{,}595 & 2{,}558 & 0 & 12 & 13.6 & 2.13 & 2.13 \\
 & Spec & 95 & 100 & 85 & 90 & 2{,}886 & 1{,}822 & 933 & 53 & 18.4 & 2.16 & 2.32 \\
\addlinespace
Haiku & HTML/JS & 100 & 100 & 93 & 93 & 5{,}538 & 2{,}926 & 2{,}454 & 56 & 21.6 & 0.20 & 0.20 \\
 & Spec & 85 & 97 & 72 & 83 & 6{,}504 & 1{,}706 & 4{,}710 & 60 & 28.0 & 0.26 & 0.30 \\
\bottomrule
\end{tabular}

%% file: tables/tests.tex
\begin{tabular}{@{}lllll@{}}
\toprule
Measure & Effect & Opus & Sonnet & Haiku \\
\midrule
H1: total output tokens & median ratio Spec/HTML & 0.81 [0.76, 0.93] & 0.99 [0.89, 1.16] & 1.13 [0.96, 1.70] \\
 & Holm-adjusted $p$ & 0.040 & 1.000 & 1.000 \\
\addlinespace
H1: generation latency & median ratio Spec/HTML & 0.90 [0.80, 1.01] & 1.23 [0.96, 1.37] & 1.25 [1.04, 1.87] \\
 & Holm-adjusted $p$ & 0.368 & 1.000 & 1.000 \\
\addlinespace
H2: valid (with repair) & mean difference (pp) & +0 [+0, +0] & +0 [+0, +0] & $-$3 [$-$8, +0] \\
 & Holm-adjusted $p$ & -- & -- & 1.000 \\
\addlinespace
H3: all tests pass (with repair) & mean difference (pp) & $-$5 [$-$10, +0] & $-$5 [$-$18, +7] & $-$10 [$-$23, +0] \\
 & Holm-adjusted $p$ & 0.750 & 0.750 & 0.750 \\
\addlinespace
Secondary: answer tokens & median ratio Spec/HTML & 0.70 [0.64, 0.81] & 0.75 [0.57, 0.85] & 0.60 [0.51, 0.67] \\
 & Holm-adjusted $p$ & $<$0.001 & $<$0.001 & $<$0.001 \\
\bottomrule
\end{tabular}

%% file: tables/taxonomy.tex
\begin{tabular}{@{}lrr@{}}
\toprule
Category & HTML/JS & Spec \\
\midrule
Wrong content (wrong algorithm step, early stop, wrong update) & 1 (1) & 4 (4) \\
Still invalid after the repair round & 0 (0) & 2 (2) \\
Extra leading state with index 0 (brief: index starts at 1) & 0 (0) & 8 (2) \\
Binary search: \texttt{mid = null} between probes (brief: only before the first) & 3 (0) & 0 (0) \\
Go-Back-N: \texttt{sent} grows within a round (ambiguity in the brief) & 3 (0) & 2 (2) \\
Harness asymmetry (runtime error swallowed, no repair; \cref{sec:res-robust}) & 0 (0) & 3 (0) \\
\midrule
Total (of 180 jobs each) & 7 (1) & 19 (10) \\
\bottomrule
\end{tabular}

%% file: tables/sensitivity.tex
\begin{tabular}{@{}lccc@{}}
\toprule
Analysis & Opus & Sonnet & Haiku \\
\midrule
Primary (frozen protocol) & 100 vs 95 & 95 vs 90 & 93 vs 83 \\
Runtime errors treated as exceptions (corrected) & 100 vs 97 & 95 vs 93 & 93 vs 83 \\
Corrected, with relaxed checkers (exploratory) & 100 vs 98 & 98 vs 97 & 100 vs 88 \\
\bottomrule
\end{tabular}

%% file: tables/usability.tex
\begin{tabular}{@{}lr@{}}
\toprule
Error raised by the runtime or validator & Artifacts \\
\midrule
Object indexed with a dynamic key / member access on a non-object & 8 \\
Array function applied to a non-array (e.g. a null initial state) & 4 \\
\texttt{=} prefix inside a plain expression field (\texttt{dur}, \texttt{derived}) & 4 \\
Malformed expression (unbalanced brackets, unquoted text) & 3 \\
Non-terminating \texttt{while} loop & 2 \\
Derived variable used inside a step & 2 \\
Other (invalid JSON, JavaScript spread syntax, non-string expression) & 3 \\
\midrule
Total spec artifacts with an error (all attempts) & 26 \\
\bottomrule
\end{tabular}

%% file: tables/qwen_descriptive.tex
\begin{tabular}{@{}llrrrrrrrr@{}}
\toprule
 & & \multicolumn{2}{c}{Valid (\%)} & \multicolumn{2}{c}{All tests pass (\%)} & \multicolumn{3}{c}{Output tokens (median)} & Latency \\
\cmidrule(lr){3-4}\cmidrule(lr){5-6}\cmidrule(lr){7-9}
Mode & Format & 1st & repair & 1st & repair & total & answer & reasoning & (s) \\
\midrule
Thinking & HTML/JS & 85 & 95 & 60 & 62 & 4{,}655 & 1{,}559 & 3{,}370 & 103.7 \\
 & Spec & 13 & 33 & 8 & 17 & 4{,}513 & 906 & 3{,}686 & 95.0 \\
\addlinespace
No thinking & HTML/JS & 83 & 97 & 35 & 40 & 1{,}618 & 1{,}618 & 0 & 29.2 \\
 & Spec & 10 & 32 & 0 & 3 & 982 & 982 & 0 & 18.1 \\
\bottomrule
\end{tabular}

%% file: tables/qwen_tests.tex
\begin{tabular}{@{}lllrr@{}}
\toprule
Test & Direction & Effect [95\% CI] & $p$ & Holm $p$ \\
\midrule
Q1 (primary): reasoning tokens & Spec $>$ HTML/JS & 1.07 [0.95, 1.33] & 0.071 & 0.286 \\
Q2a: answer tokens & Spec $<$ HTML/JS & 0.52 [0.49, 0.61] & $<$0.001 & $<$0.001 \\
Q2b (H1): total output tokens & Spec $<$ HTML/JS & 0.87 [0.76, 1.04] & 0.184 & 0.552 \\
Q2c (H1): generation latency & Spec $<$ HTML/JS & 0.89 [0.77, 1.11] & 0.324 & 0.648 \\
Q3 (H2): valid (with repair), pp & Spec $>$ HTML/JS & $-$62 [$-$78, $-$43] & 1.000 & 1.000 \\
Q4 (H3): all tests pass (with repair), pp & two-sided & $-$45 [$-$63, $-$27] & $<$0.001 & 0.004 \\
\midrule
\multicolumn{5}{@{}l}{\emph{No-thinking ablation (exploratory): gain in all tests pass from thinking, pp}} \\
HTML/JS & two-sided & +22 [+2, +42] & 0.098 & \\
Spec & two-sided & +13 [+0, +30] & 0.188 & \\
Spec minus HTML/JS & two-sided & $-$8 [$-$32, +17] & 0.609 & \\
\bottomrule
\end{tabular}